\documentclass[a4paper,fleqn,3p]{elsarticle}

\usepackage[utf8]{inputenc}
\usepackage[T1]{fontenc}
\usepackage{textcomp}
\usepackage[english]{babel}
\usepackage{microtype} 
\usepackage{comment}
\usepackage{breakcites}
\usepackage{graphicx}
\usepackage{adjustbox}
\usepackage{tabulary}
\usepackage{ltablex}
\usepackage{booktabs}
\usepackage{float}
\usepackage[colorinlistoftodos]{todonotes}
\usepackage{multicol}
\usepackage{amssymb}
\usepackage{enumitem}
\usepackage{array}
\usepackage{tikz}
\usepackage{algorithm}
\usepackage{algorithmic}
\usepackage{amsmath}
\usepackage[edges]{forest}
\usepackage[colorlinks=true, linkcolor=black,citecolor=black]{hyperref}

\journal{Knowledge-Based Systems}

\newcolumntype{s}{>{\hsize=.5\hsize}X}
\newcolumntype{t}{>{\hsize=.25\hsize}X}

\usetikzlibrary{shapes.geometric,arrows.meta}

\definecolor{Gray}{gray}{0.9}
\definecolor{LightCyan}{rgb}{0.88,1,1}

\def\tsc#1{\csdef{#1}{\textsc{\lowercase{#1}}\xspace}}
\tsc{WGM}
\tsc{QE}
\tsc{EP}
\tsc{PMS}
\tsc{BEC}
\tsc{DE}
\definecolor{myorange}{RGB}{245, 121, 58}
\definecolor{mypurple}{RGB}{169, 90, 161}
\definecolor{mylightblue}{RGB}{133, 192, 249}
\definecolor{myblue}{RGB}{15, 32, 128}
\newcommand{\citeColored}[2]{\hypersetup{citecolor=#1}\cite{#2}\hypersetup{citecolor=black}}

\usepackage{etoolbox}
\makeatletter
\patchcmd{\ps@pprintTitle}% <cmd>
  {Preprint submitted to}% <search>
  {Article accepted at}% <replace>
  {}{}% <succes><failure>
\makeatother

\begin{document}

\title{Explainability in Deep Reinforcement Learning}

\author[1]{Alexandre Heuillet\fnref{fn1}}
\ead{aheuillet@enseirb-matmeca.fr}

\address[1]{ENSEIRB-MATMECA, Bordeaux INP, 1 avenue du Docteur Albert Schweitzer, 33400 Talence, France}

\author[2]{Fabien Couthouis\fnref{fn1}}
\ead{fcouthouis@ensc.fr}
\address[2]{ENSC, Bordeaux INP, 109 avenue Roul, 33400 Talence, France}

\author[3]{Natalia Díaz-Rodríguez\corref{cor1}}
\ead{natalia.diaz@ensta-paris.fr}
\address[3]{ENSTA Paris, Institut Polytechnique Paris, Inria Flowers Team, 828 boulevard des Maréchaux, 91762 Palaiseau, France}

\cortext[cor1]{Corresponding author}
\fntext[fn1]{Equal contribution}

\begin{abstract}
    A large set of the explainable Artificial Intelligence (XAI) literature is emerging on feature relevance techniques to explain a deep neural network (DNN) output or explaining models that ingest image source data.
    However, assessing how XAI techniques can help understand models beyond classification tasks, e.g. for reinforcement learning (RL), has not been extensively studied. We review recent works in the direction to attain Explainable Reinforcement Learning (XRL), a relatively new subfield of Explainable Artificial Intelligence, intended to be used in general public applications, with diverse audiences, requiring ethical, responsible and trustable algorithms. In critical situations where it is essential to justify and explain the agent's behaviour, better explainability and interpretability of RL models could help gain scientific insight on the inner workings of what is still considered a black box. We evaluate mainly studies directly linking explainability to RL, and split these into two categories according to the way the explanations are generated: transparent algorithms and post-hoc explainability. We also review the most prominent XAI works
    from the lenses of how they could potentially enlighten the further deployment of the latest advances in RL, in the demanding present and future of everyday problems. 
\end{abstract}


\maketitle

\section{Introduction} \label{sec:intro}
 
During the past decade, Artificial Intelligence (AI), and by extension Machine Learning (ML), have seen an unprecedented rise in both industry and research. The progressive improvement of computer hardware associated with the need to process larger and larger amounts of data made these underestimated techniques shine under a new light. Reinforcement Learning (RL) focuses on learning how to map situations to actions, in order to maximize a numerical reward signal \cite{Sutton_Barto}. The learner is not told which actions to take, but instead must discover which actions are the most rewarding by trying them. %In more technical terms, r
Reinforcement learning addresses the problem of how agents should learn a policy that take actions to maximize the cumulative reward through interaction with the environment \cite{duan2016benchmarking}. 

Recent progress in Deep Learning (DL) for learning feature representations has significantly impacted RL, and the combination of both methods (known as deep RL) has led to remarkable results in a lot of areas. Typically, RL is used to solve optimization problems when the system has a very large number of states and has a complex stochastic structure. Notable examples include training agents to play Atari games based on raw pixels \cite{mnih2013playing,mnih2015human}, board games \cite{silver2017mastering, Silver2017MasteringTG}, complex real-world robotics problems such as manipulation \cite{Andrychowicz_2019} or grasping \cite{kalashnikov2018qtopt} and other real-world applications such as resource management in computer clusters \cite{Mao:2016}, network traffic signal control \cite{rl_traffic_signal}, chemical reactions optimization \cite{doi:10.1021/acscentsci.7b00492} or recommendation systems \cite{zheng_news_recommendation}.

The success of Deep RL could augur an imminent arrival in the industrial world. However, like many Machine Learning algorithms, RL algorithms suffer from a lack of explainability. This defect can be highly crippling as many promising RL applications (defense, finance, medicine, etc.) need a model that can explain its decisions and actions to human users \cite{gunning2019darpa} as a condition to their full acceptation by society.
Furthermore, deep RL models are complex to debug for developers, as they rely on many factors: environment (in particular the design of the reward function), observations encoding, large DL models and the algorithm used to train the policy. Thus, an explainable model could aid fixing problems quicker and drastically speed up new development in RL methods. Those last two points are the main arguments in favor of the necessity of explainable reinforcement learning (XRL).

While explainability starts being well developed for standard ML models and neural networks \cite{ribeiro2016i, lundberg2017unified, Selvaraju_2019}, the particular domain of RL has yet many intricacies to be better understood: both in terms of its functioning, and in terms of conveying the decisions of an RL model to different audiences. The difficulty lies in the very recent human-level performance of deep RL algorithms and by their complexity, normally parameterized with thousands if not millions of parameters \cite{Brown20}. The present work intends to provide a non-exhaustive state-of-the-art review on explainable reinforcement learning, highlighting the main methods that 
we envision most promising. In the following, we will briefly recall some important concepts in XAI.

\subsection{Explainable AI: Audience \label{sect:audience}}

Explaining a Machine Learning model may involve different goals: trustworthiness, causality, transferability, informativeness, fairness, confidence, accessibility, interactivity and privacy awareness. These goals have to be taken into account while explaining a model because the expected type of explanation may differ, depending on the pursued objective. For example, a saliency map explaining what is recognized as a dog on an input image does not tell us much about privacy awareness. In addition, each goal may be a dimension of interest, but only for a certain audience (the public to whom the explanations will be addressed). Indeed, the transferability of a model can be significative for a developer, since he/she can save time by training only one model for different tasks, while the user will not be impacted, if not aware, by this aspect.
 
The understandability of an ML model therefore depends on its transparency (its capacity to be understandable by itself) but also on human understanding. According to these considerations, it is essential to take into account the concept of \textit{audience}, as the intelligibility and comprehensibility of a model is dependant on the goals and the cognitive skills of its users. Barredo Arrieta et al. \cite{Barredo20} discuss these aspects with additional details.

\begin{tabularx}{\textwidth}{s X X X}
     \caption{Target audience in XAI. This table shows the different objectives of explainability in Machine Learning models for different audience profiles. Inspired from the diagram presented in Barredo Arrieta et al. \cite{Barredo20}.}\\
     \hline
     \textbf{Target audience} & \textbf{Description} & \textbf{Explainability purposes} & \textbf{Pursued goals}\\ 
     \hline
        Experts & Domain experts, model users (e.g. medical doctors, insurance agents) &  Trust the model itself, gain scientific knowledge & Trustworthiness, causality, transferability, informativeness, confidence, interactivity\\
        \hline
        Users & Users affected by model decisions & Understand their situation, verify fair decisions. & Trustworthiness, informativeness, fairness, accessibility, interactivity, privacy awareness\\
        \hline
        Developers & Developers, researchers, data scientists, product owners... & Ensure and improve product efficiency, research, new functionalities... & Transferability, informativeness, confidence\\
        \hline
        Executives & Managers, executive board members... &  Assess regulatory compliance, understand corporate AI applications... & Causality, informativeness, confidence\\
        \hline
        Regulation &  Regulatory entities/agencies & Certify model compliance with the legislation in force, audits, ... & Causality, informativeness, confidence, fairness, privacy awareness\\
        \hline
    \label{tab:audience}
\end{tabularx}

\subsection{Evaluating explanations }

The broad concept of evaluation is based on metrics aiming to compare how well a technique performs compared to another. In the case of model explainability, metrics should evaluate how well a model fits the definition of explainable and how well performs in a certain aspect of explainability.

Explanation evaluation in XAI has proven to be quite a challenging task. First because the concept of explainability in Machine Learning is not well or uniformly accepted by the community and there is not a clear definition and thus, not a clear consensus on which metrics to use. Secondly because an explanation is relative to a specific audience, which is sometimes difficult to deal with (in particular when this specific audience is composed of domain experts who can be hard to involve in a testing phase). Thirdly, because the quality of an explanation is always qualitative and subjective, since it depends on the audience, the pursued goal and even the human variability as two people can have a different level of understandability for the same explanation. That is why user studies are so popular to evaluate explanations as it makes possible to convert qualitative evaluations into quantitative ones, by asking questions on the accuracy and clarity of the explanation such as "Does this explanation allow you to understand why the model predicted that this image is a dog? Did the context helped the model?"... etc. 
Generally in XAI, there is only a single model to explain at a time; however, it is more complicated in XRL, as we generally want to explain a policy, or "Why the agent took action x in state s?".

Doshi-Velez et al. \cite{doshivelez2017rigorous} propose an attempt to formulate some approaches to evaluate XAI methods. The authors introduce three main levels to evaluate the quality of the explanations provided by an XAI method, as summarized in Table \ref{tab:explanation_levels}.

\begin{tabularx}{\linewidth}{t t t X}
        \caption{Three main levels for evaluating the explanations provided by an XAI method (inspired from explanations provided in Doshi-Velez et al. \cite{doshivelez2017rigorous}).}\\
        \hline
        \textbf{Level of evaluation} & \textbf{Type of task} & \textbf{Required humans} & \textbf{Modus Operandi} \\
        \hline
        Application level & Real task & Domain expert &  Put the explanation into the product and have it tested by the end user. \\
        \hline
        Human level & Simplified task & Layperson & Carry out the application level experiments with laypersons as it makes the experiment cheaper and it is easier to find more testers. \\
        \hline
        Function level & Proxy task & No human required &  Uses a proxy to evaluate the explanation quality. Works best when the model class used has already been evaluated by someone else in human level. For instance, a proxy for decision trees can be the depth of the tree.\\
        \hline
        \label{tab:explanation_levels}
\end{tabularx}

A common example of evaluation of an application level or human level task is to evaluate the quality of the mental model built by the user after seeing the explanation(s). Mental models can be described as internal representations, built upon experiences, and which allow to mentally simulate how something works in the real world. Hoffman et al. \cite{hoffman2018metrics} propose to evaluate mental models by 1. Asking post-task questions on the behavior of the agent (such as "How does it work?" or "What does it achieve?") and 2. Asking the participants to make predictions on the agent's next action. These evaluations are often done using Likert scales.

\subsection{Organisation of this work \label{subsect:organisation}}
In this survey, we first introduce XAI and its main challenges in Section \ref{sec:intro}. We then review the recent literature in XAI applied to reinforcement learning in Section \ref{sec:review}. In Section \ref{sec:discussion}, we discuss the different approaches employed in the literature. Finally, we conclude in Section \ref{sec:conclusion} with some directions for future research.

The key contributions of this paper are as follows:
\begin{itemize}
    \item A recent state of the art for Explainable Reinforcement Learning.
    \item An attempt to categorize XRL methods.
    \item Discussion and future work recommendations (provided at the end, in Sections \ref{sec:discussion} and \ref{sec:conclusion}). 
\end{itemize}

We hope that this work will give more visibility to existing XRL methods, while helping developing new ideas in this field.

\section{XAI in RL: State of the art and reviewed literature}
\label{sec:review}
We reviewed the state of the art on XRL and summarized it in Table \ref{tab:summary_papers}. This table presents, for each paper, the task(s) for which an explanation is provided, the employed RL algorithms (whose Algorithms glossary can be found in the \ref{ref:appendix}), and the provided type of explanations, i.e.: based on images, diagrams (graphical components such as bar charts, plots or graphs), or text.
We also present the level of the provided explanation (\textit{local} if it explains only predictions, \textit{global} if it explains the whole model), and the audience concerned by the explanation, as discussed in Section \ref{sect:audience}.

\begin{tabularx}{\linewidth}
    {|>{\raggedright\arraybackslash}X
     |>{\raggedright\arraybackslash}s
     |>{\raggedright\arraybackslash}s
     |>{\raggedright\arraybackslash}s
     |>{\raggedright\arraybackslash}s
     |>{\raggedright\arraybackslash}s
     |>{\raggedright\arraybackslash}s|}
    \caption{Summary of reviewed literature on explainable RL (XRL) and deep RL (DRL).}
    \label{tab:summary_papers}\\
    \hline\endfirsthead
    \hline\multicolumn{6}{c}{\itshape continues on next page}\endfoot
    \endlastfoot
    \textbf{Reference}&\textbf{Task/ Environment}&\textbf{Decision process} &\textbf{Algorithm(s)}&\textbf{Explanation type (Level)} &\textbf{Target} \\
    \hline
    
    % %Transparent
     Relational Deep RL \cite{zambaldi2018relational} & Planning + strategy games (Box-World/ Starcraft II) & POMDP & IMPALA & Images (Local) & Experts\\
     \hline
     Symbolic RL with Common Sense \cite{2018arXiv180408597D} & Game (object retrieval) & POMDP & SRL+CS, DQL & Images (Global) & Experts \\
     \hline
     Decoupling feature extraction from policy learning \cite{Raffin19} & Robotics (grasping), and navigation & MDP & PPO & Diagram (state plot \& image slider (Local)
     & Experts \\
    \hline
    Explainable RL via Reward Decomposition \cite{juozapaitisexplainable} & Game (grid and landing) & MDP & HRA, SARSA, Q-learning & Diagrams (Local) & Experts, Users, Executives \\
    \hline
    Explainable RL Through a Causal Lens \cite{madumal2019explainable} & Games (OpenAI benchmark and Starcraft II) & Both & PG, DQN, DDPG, A2C, SARSA & Diagrams, Text (Local) & Experts, Users, Executives \\
    \hline
    Shapley Q-value: A Local Reward Approach to Solve Global Reward Games \cite{wang2019shapley} & Multiagents (Cooperative Navigation, Prey-and-Predator and Traffic Junction) & POMDP & DDPG & Diagrams (Local) & Experts \\
    \hline
    Dot-to-Dot: Explainable HRL For Robotic Manipulation \cite{Dot-to-Dot} & Robotics (grasping) & MDP & DDPG, HER, HRL & Diagrams (Global) & Experts, Developers\\
    \hline
    Self-Educated Language Agent With HER For Instruction Following \cite{cideron2019selfeducated} & Instruction Following (MiniGrid) & MDP & Textual HER & Text (Local) & Experts, Users, Developers\\
    \hline
     Commonsense and Semantic-guided Navigation
     \cite{yucommonsense} & Room navigation & POMDP & - & Text (Global) & Experts \\
    \hline
     %A 
     Boolean Task Algebra
     \cite{tasse2020boolean} & Game (grid) & MDP & DQN & Diagrams & Experts\\
    \hline
    %Post-Hoc
     Visualizing and Understanding Atari
     \cite{atarivisualizing} & Games (Pong, Breakout, Space Invaders) & MDP & A3C & Images (Global) & Experts, Users, Developers \\
    \hline
    Interestingness Elements for XRL through Introspection \cite{Sequeira2019InterestingnessEF, sequeira2019interestingness2} & Arcade game (Frogger) & POMDP & Q-Learning & Images (Local) & Users \\
    \hline
    %Other
     Composable DRL for Robotic Manipulation \cite{Haarnoja_2018} & Robotics (pushing and reaching) & MDP & Soft Q-learning & Diagrams (Local) & Experts \\
     \hline
     Symbolic-Based Recognition of Contact States for Learning Assembly Skills \cite{10.3389/frobt.2019.00099} & Robotic grasping & POMDP & HMM, PAA, K-means & Diagrams (Local) & Experts\\
    \hline
    Safe Reinforcement Learning with Model Uncertainty Estimates \cite{ltjens2018safe} & Collision avoidance & POMDP & Monte Carlo Dropout, bootstrapping & Diagrams (Local) & Experts\\
    \hline
\end{tabularx}

\begin{figure}[H]
    \centering
 \label{fig:taxonomic_tree}
 \begin{forest}
    for tree={
      grow=east,
      parent anchor=south east,
      child anchor=south west,
      anchor=south,
      align=center,
      l sep+=2.5pt,
      s sep+=-5pt,
      inner sep=0pt,
      outer sep=0pt,
      edge path={
        \noexpand\path [draw, rounded corners=5pt, \forestoption{edge}] (!u.parent anchor) [out=0, in=180] to (.child anchor)\forestoption{edge label} -- (.south east);
      },
      for root={
        ellipse,
        draw,
        parent anchor=east,
      },
    }
    [XAI in\\RL
    [
      [Post-hoc\\explainability
        [Saliency maps
          [\citeColored{myorange}{atarivisualizing}
          ]
        ]
        [Interaction data
          [\citeColored{myorange}{Sequeira2019InterestingnessEF} \citeColored{myorange}{sequeira2019interestingness2}
        ]
      ]
      ]
      [Transparent\\algorithms
        [Hierarchical learning
          [\citeColored{mypurple}{cideron2019selfeducated}  \citeColored{mypurple}{yucommonsense} \citeColored{mylightblue}{Dot-to-Dot} \citeColored{mylightblue}{tasse2020boolean}
          ]
        ]
        [Simultaneous leaning
         [\citeColored{mylightblue}{juozapaitisexplainable} \citeColored{mypurple}{madumal2019explainable}
         \citeColored{mylightblue}{wang2019shapley} \citeColored{mylightblue}{ltjens2018safe}
          ]
        ]
        [Representation learning
            [\citeColored{myorange}{zambaldi2018relational} \citeColored{myorange}{2018arXiv180408597D} \citeColored{mylightblue}{Raffin19} \citeColored{mylightblue}{raffin2018srl} \\ \citeColored{mylightblue}{garnelo2019} \citeColored{myorange}{2018arXiv180408597D}
            \citeColored{mylightblue}{pathak2017curiositydriven}
          ]
        ]
      ]
    ]
    ]
\end{forest}
\caption{
Taxonomy of the reviewed literature identified for bringing explainability to RL models. References in \textcolor{myorange}{orange}, \textcolor{mypurple}{purple}, and \textcolor{mylightblue}{light blue} correspond to XAI techniques using \textcolor{myorange}{images}, \textcolor{mypurple}{text} or \textcolor{mylightblue}{diagrams}, respectively.
}
\end{figure}
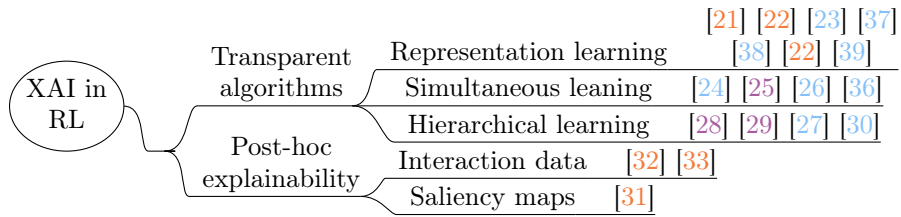

\label{tab:guidelines}
 In Table \ref{tab:summary_papers} we summarized the literature focusing on explainable fundamental RL algorithms. However, we also reviewed articles about state of the art XAI techniques that can be used in the context of current RL which we did not include in Table \ref{tab:summary_papers}. Next, we will describe the main ideas provided by these papers which can help bring explainability in RL. It is possible to classify all recent studies in two main categories: transparent methods and Post-Hoc explainability according to the XAI taxonomies in Barredo Arrieta et al. \cite{Barredo20}. On the one hand, inherently transparent algorithms include by definition every algorithm which is understandable by itself, such as a decision-trees. On the other hand, Post-Hoc explainability includes all methods that provide explanations of an RL algorithm after its training, such as SHAP (SHapley Additive exPlanations) \cite{lundberg2017unified} or LIME \cite{ribeiro2016i} for standard ML models. Reviewed papers are referenced by type of explanation in Figure \ref{fig:taxonomic_tree}.

%Taxonomic tree

\subsection{Transparent algorithms}
 
Transparent algorithms are well known and used in standard Machine Learning (e.g., linear regression, decision trees or rule-based systems). Their strength lie in the fact that they are designed to have a transparent architecture that makes them explainable by themselves, without the need of any external processing. However, it is quite different for RL, as standard DRL algorithms (e.g., DQN, PPO, DDPG, A2C...) are not transparent by nature. In addition, the large majority of studies related to transparency in XRL chose to build algorithms targeting only a specific task. Nonetheless, most of the time and contrary to standard Machine Learning models, transparent RL algorithms can achieve state of the art performance in these specific tasks \cite{wang2019shapley, 2018arXiv180408597D, zambaldi2018relational}.

\subsubsection{Explanation through representation learning}
\label{subsubsec:replearn}
 
Representation learning algorithms focuses on learning abstract features that characterize data, in order to make it easier to extract useful information when building predictors \cite{lesort2018srlOverview, bengio2012representation}. These learned features have the advantage of having low dimensionality, which generally improves training speed and generalization of Deep Learning models
\cite{Lesort19RoboticPriors, lesort2018srlOverview, Raffin19}. 

In the context of RL, learning representations of states, actions or policy can be useful to explain a RL algorithm, as these representations can bring some clues on the functioning of the algorithm. 
Indeed, State Representation Learning (SRL) \cite{lesort2018srlOverview} is a particular type of representation learning that aims at building a low-dimensional and meaningful representation of a state space, by processing  high-dimensional raw observation data (e.g., learn a position (x, y) from raw image pixels). This enables to capture the variations in the environment influenced by the agent’s actions and thus, extrapolate explanations. SRL can be especially useful in RL for robotics and control \cite{raffin2018srl, Raffin19, traor2019discorl, doncieux2020dream, doncieuxSRL2018},
and can help to understand how the agent interprets the observations and what is relevant to learn to act, i.e., actionable or controllable features \cite{Lesort19RoboticPriors}. Indeed, the dimensionality reduction induced by SRL, coupled with the link to the control and possible disentanglement of variation factors, could be highly beneficial to improve our understanding capacity of the decisions made by RL algorithms using a state representation method \cite{lesort2018srlOverview}.
For example, SRL can be used to split the state representation \cite{Raffin19} 
according to the different training objectives to be optimized before learning a policy. This allows to allocate room for encoding each necessary objective within the embedding state to be learned (in that case, reward prediction, a reconstruction objective and an inverse model). In this context, tools such as S-RL Toolbox  \cite{raffin2018srl}
allow sampling from the embedding state space (learned through SRL) to allow a visual interpretation of the model's internal state, and pairing it with its associated input observation. Comprehensibility is thus enhanced,
more easily observing if smoothness is preserved in the state space, as well as whether other invariants related to learning specific control task are guaranteed. 

There are several approaches employed for SRL: reconstructing the observations using autoencoders \cite{alvernaz2017autoencoderaugmented, finn2015deep}, training a forward model to predict next state \cite{Hoof2016StableRL, pathak2017curiositydriven}, teach to an inverse model how to predict actions from previous state(s) \cite{shelhamer2016loss, pathak2017curiositydriven} or using prior knowledge to constrain the state space \cite{jonschkowski2015,Lesort19RoboticPriors}. 

Along the same lines, learning disentangled representations
\cite{higgins2018definition, AchilleNIPS2018_8193, aless2017emergence, CasellesNIPS2019_8709}
is another interesting idea used for unsupervised learning, which decomposes (or disentangles) each feature into narrowly defined variables and encodes them as separate low-dimensional features (generally using a Variational Autoencoders \cite{kingma2013autoencoding}).
It is also possible to make use of this concept, as well as lifelong learning to learn more interpretable representations on unsupervised classification tasks. In addition, one could argue that learning through life would allow compacting and updating old knowledge with new one while preventing catastrophic forgetting \cite{Lesort2019ContinualLF}. Thus, this is a key concept that could lead to more versatile RL agents, being able to learn new tasks without forgetting the previous ones.
Information Maximizing Generative Adversarial Networks (InfoGAN) \cite{InfoGAN} is another model based on the principles of learning disentangled representations. The noise vector used in traditional GANs is decomposed into two parts: $z$: incompressible noise; and $c$: the latent code used to target the salient semantic features of the data distribution. The main idea is to feed $z$ and $c$ to the generator $G$, to maximize the mutual information between $c$ and $G(z, c)$, in order to assure that the information contained in $c$ 
is preserved during the generation process. As a result, the InfoGAN model is able to create an interpretable representation via the latent code $c$ (i.e., the
values changing according to shape and features of the input data).

Some work has been done to learn representations by combining symbolic AI with deep RL in order to facilitate the use of background knowledge, the exploitation of learnt knowledge, and to improve generalization \cite{garnelo2019, garcez2015, santoro2017simple, garnelo2016deep}. 
Consequently, it also improves the explainability of the algorithms, while preserving state-of-the-art performance.

\begin{figure}[htbp!]
    
    \centering
    \includegraphics[width=\textwidth]{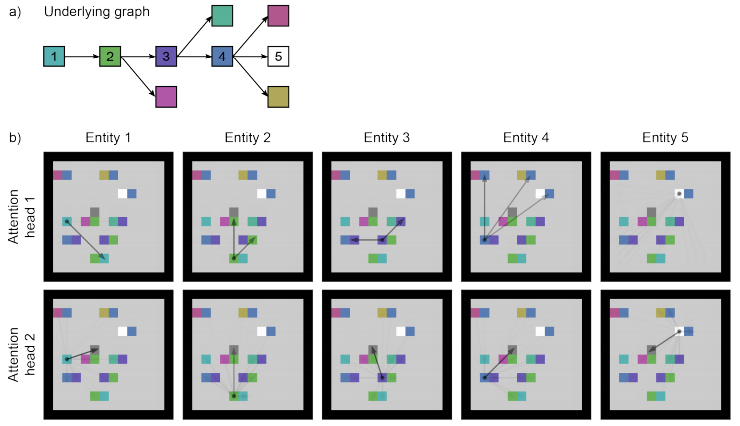}
    \caption{
    Visualization of attention weights for the \textit{Box-World} task 
    (environment created by authors of \cite{zambaldi2018relational}), in which the agent has to open boxes to obtain either keys or rewards. 
\textbf{(a)} The underlying graph of one example level.
\textbf{(b)} The result of the analysis for that level, using each entity (represented as colored pixels) along the solution path (1–5) as the source of attention. Boxes are represented by two adjacent colored pixels. On each box, the pixel on the right represents the box’s lock and its color indicates which key can be used to open it. The pixel on the left indicates the content of the box which is inaccessible while
the box is locked.  Arrows point to the entities that the source is attending to. The arrow’s transparency is determined by the corresponding attention weight (reproduced with permission of Vinicius Zambaldi \cite{zambaldi2018relational}).}
\label{fig:boxworld}

\end{figure}

Zambaldi et al. \cite{zambaldi2018relational} propose making use of Inductive Logic Programming and self-attention to represent states, actions and policies using first order logic, using a mechanism similar to graph neural networks and more generally, message passing computations \cite{denil2017programmable, kipf2016semisupervised, battaglia2018relational, 4700287}.
In these kind of models entity-entity relations are explicitly computed when considering the messages passed between connected nodes of the graph as shown in Fig. \ref{fig:boxworld}. Self-attention is used here as a method to compute interactions between these different entities (i.e. relevant pixels in a RGB image for the example from \cite{zambaldi2018relational}), and thus perform non-local pairwise relational computations.
This technique allows an expert to visualize the agent's attention weights associated to its available actions and interpret how to improve the understanding of its strategy. 

Another work that aims to incorporate common sense to the agent, in terms of symbolic abstraction to represent the problem, is in \cite{2018arXiv180408597D}. 
This method subdivides the world state representation into many sub-states, with a degree of associated importance based on how far the object is from the agent. This helps understand the relevance of the actions taken by the agent by determining which sub-states were chosen.

\subsubsection{Simultaneous learning of the explanation and the policy}
 
\begin{figure}[htbp!]
    \centering
    \includegraphics[width=\textwidth]{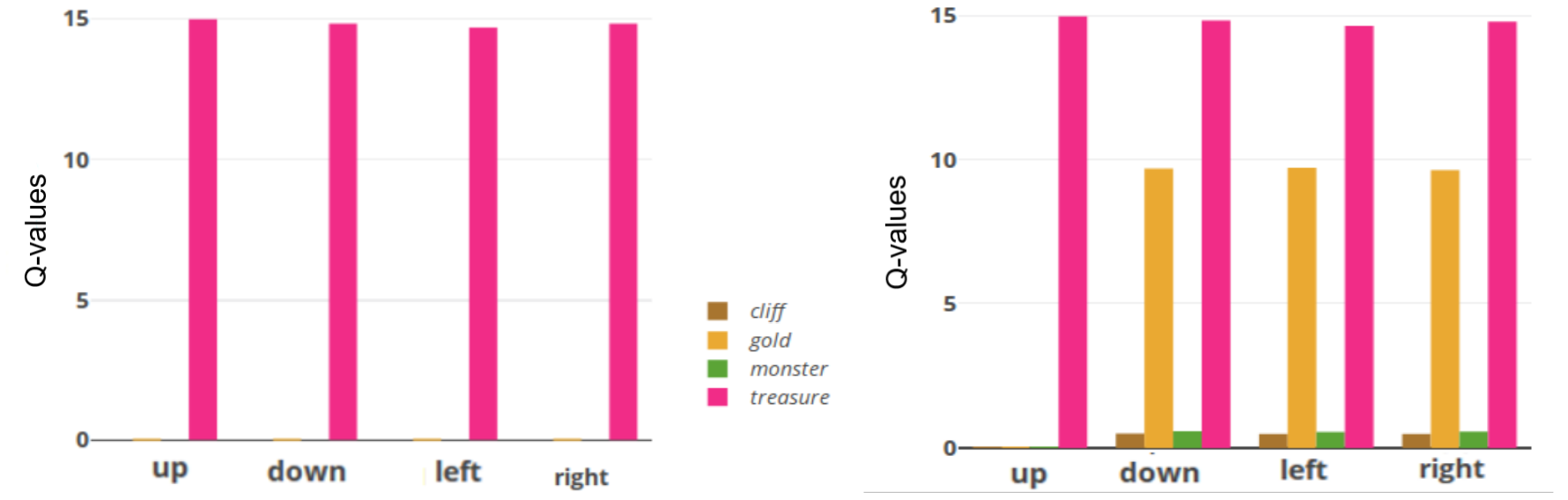}
    \caption{\textbf{Left} Reward Decompositions for DQN.
    \textbf{Right} Hybrid Reward Architecture (HRA) at cell (3,4) in Cliffworld. HRA predicts an extra “gold” reward for actions which do not lead to a terminal state. Reproduced with permission of Zoe Juozapaitis \cite{juozapaitisexplainable}.}
    \label{fig:RDX}
\end{figure}

\begin{figure}[htbp!]
    \centering
    \includegraphics[scale=0.65]{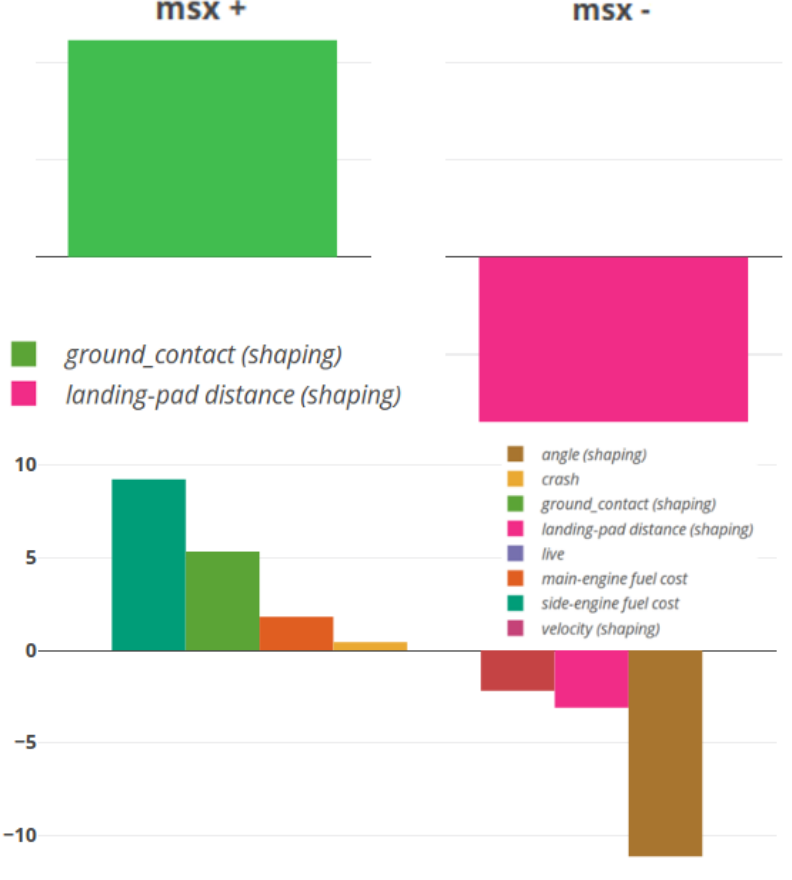}
    \caption{\textbf{Top} Minimal Sufficient Explanations (MSX) (fire  down engine action
    vs. \textit{do nothing} action) for decomposed reward DQN in \textit{Lunar Lander} environment near landing site. The shaping rewards dominate decisions. \textbf{Bottom} RDX (noop vs. fire-main-engine) for HRA in Lunar Lander before a crash. The RDX shows that noop is preferred to avoid penalties such as fuel cost. Reproduced with permission of Zoe Juozapaitis \cite{juozapaitisexplainable}.}
    \label{fig:MSX}
\end{figure}

While standard DRL algorithms struggle to provide explanations, those can be tweaked to learn simultaneously both policy and explanation. Thus, explanations become a learned component of the model. These methods are recommended on specific problems where it is possible to introduce knowledge, such classifying rewards by types, adding relationships between states, etc...
Thus, tweaking the algorithm to introduce some task knowledge and to learn explanations generally also improves performance. A general notion is that the knowledge gained from the auxiliary task objective must be useful for downstream tasks.
In this direction, Juozapaitis et al. \cite{juozapaitisexplainable} introduced reward decomposition, whose main principle is to decompose the reward function into a sum of meaningful reward types. Authors used reward decomposition to improve performance on Cliffworld and Starcraft II, where each action can be classified according to its type. This method consists of using a custom \textit{decomposed reward DQN} by defining a vector-valued reward function, where each component 
is the reward for a certain type so that actions can be compared in terms of
trade-offs among the types. In the same way, the Q-function is also vector valued and each component
gives action values that account for only a reward type. The sum of each of those vector-valued functions gives the overall Q or reward function. Learning multiple Q-functions, one for each type of reward, allows the model to learn the best policy while also learning the explanations (i.e. the type of reward that the agent wanted to maximize by his action, illustrated on Fig. \ref{fig:MSX}).
They introduce the concept of \textit{Reward Difference Explanation} (RDX, in Fig. \ref{fig:RDX}) which enables to understand the reasons why an action has an advantage (or disadvantage) over another. They also define \textit{Minimal Sufficient Explanations} (MSX, See Fig. \ref{fig:MSX}), in order to help humans identify a small set of the most important \textit{reasons} why the agent choose specific actions over another. MSX+ and MSX- are sets of
critical positive and negative reasons (respectively) for the actions preferred by the agent.

While reward decompositions help to understand the agent choice preferences between several actions, minimal sufficient explanations are used to help selecting the most important reward decompositions. Other works that facilitate the explainability of RL models by using reward-based losses for more interpretable RL are in \cite{zhang2018decoupling, shelhamer2016loss, pathak2017curiositydriven}.

\begin{figure}[htbp!]
    \centering
    \includegraphics[scale=0.3]{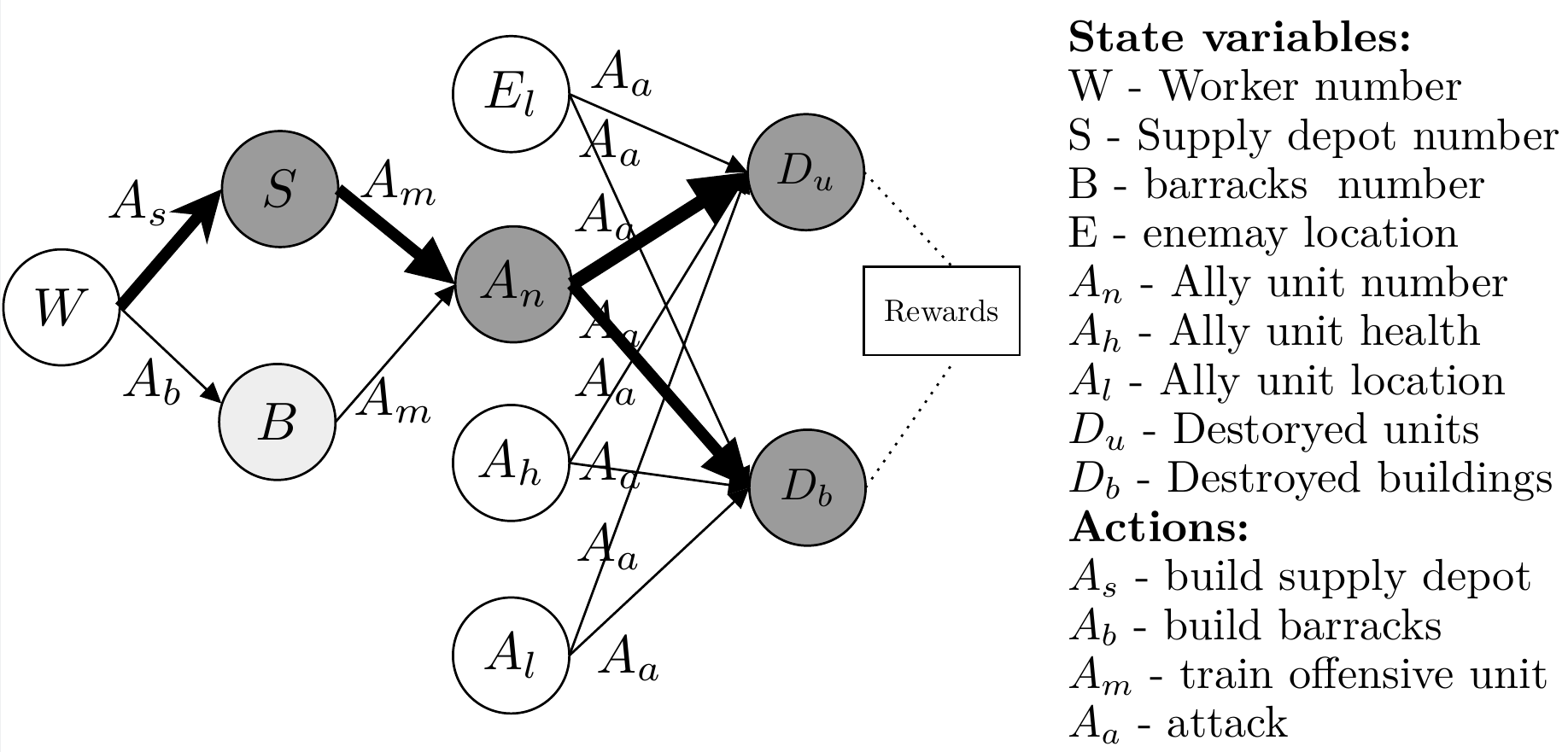}
    \caption{Action influence graph of a Starcraft II agent. The causal chain (explanation) for action $A_s$ is depicted in bold arrows and the extracted \textit{explanan} (subset of causes given the explanation) is shown as darkened nodes. The counterfactual action (\textit{why not $A_b$}) \textit{explanan} is shown as grayed node (B). Here, $A_s$ is the \textit{explanandum}, the action for which the user needs explanation. Thus, we can answer the question \textit{"Why not build\_barrack $(A_b$)?"}. Indeed, the explanation provided by the graph in bold arrows is: \textit{"Because it is more desirable to do action build\_supply\_depot ($A_s$) to have more Supply Depots as the goal is to have more Destroyed Units ($D_u$) and Destroyed Buildings ($D_b$)"}.
    Reproduced with permission of \cite{madumal2019explainable}.}
    \label{fig:influencegraph}
\end{figure}

In the same vein, Madumal et al. \cite{madumal2019explainable} use the 
way humans understand and represent knowledge through causal relationships and introduce an \textit{action influence model}: a causal model which can explain 
 the behaviour agents using causal explanations.
Structural causal models \cite{10.1093/bjps/axi147} represent the world using random variables, some of which might have causal relationships, which can be described thanks to a set of structural equations. In this work, structural causal models are extended to include actions as part of the causal relationships. An action influence model is a tuple represented by the state-actions ensemble and the corresponding set of structural equations. The whole process is divided into 3 phases:
        \begin{itemize}
            \item Defining the qualitative causal relationships of variables as an action influence model. 
            \item Learning the structural equations (as multivariate regression models during the training phase of the agent).
            \item Generating explanations, called \textit{explanans}, by traversing the action influence graph (see Figure \ref{fig:influencegraph}) from the root to the leaf reward node.
        \end{itemize}
        
This kind of models allow encoding cause-effect relations between events (actions and states) as shown by the graph featured in Figure \ref{fig:influencegraph}. Thus, they can be used to generate explanations of the agent behavior ("why" and "why not" questions), based on knowledge about how actions influence the environment. Their method was evaluated through a user study showing that, compared to video game playing without any explanations and relevant variable explanations, this model performs significantly better on 1) task prediction and 2) explanation \textit{goodness}. However, trust was not shown to be significantly improved.

Authors of \cite{ltjens2018safe} also learn explanations along with the model policy on pedestrians collision avoidance tasks. In this paper, an ensemble of LSTM networks was trained using  Monte Carlo Dropout \cite{gal2015dropout} and bootstrapping \cite{osb2016deep} to estimate collision probabilities and thus predict uncertainty estimates to detect novel observations. The magnitude of those uncertainty estimates was shown to reveal novel obstacles in a variety of scenarios, indicating that the model knows what it does not know. The result is a collision avoidance policy that can measure the novelty of an observation (via model uncertainty) and cautiously avoids pedestrians that exhibit unseen behavior.
Measures of model uncertainty can also be used to identify unseen data during training or testing. Policies during simulation demonstrated to be more robust to novel observations and take safer actions than an uncertainty-unaware baseline. This work also responds to the problem of safe reinforcement learning \cite{garcia15SafeRL}, whose goal is to ensure reasonable system performance and/or respect safety constraints %during learning and/or deployment processes.
also at the deployment phase.
 
Some work has also been made to explain multiagent RL. Wang et al. \cite{wang2019shapley} developed an approach named Shapley Q-values Deep Deterministic Policy Gradient (SQDDPG) to solve global reward games in a multiagent context based on Shapley values and DDPG. 
The proposed approach relies on distributing the global reward more efficiently across all agents. They show that integrating Shapley values into DDPG enables to share the global reward between all agents according to their contributions: the more the agent contributes, the more reward it will get. This contrasts to the classical shared reward approach, which could cause inefficient learning by assigning rewards to an agent who contributed poorly. The experiments showed that SQDDPG presents faster convergence rate and fairer credit assignment in comparison with other algorithms (i.e. IA2C, IDDPG, COMA and MADDPG). This method allows to plot credit assignment to each agent, which can explain how the global reward is divided during training and what agent contributed the most to obtain the global reward.
%Size
   \begin{figure}[htbp!]
    \centering
    \includegraphics[scale=1]{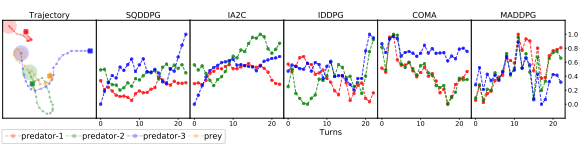}
    \caption{Credit assignment to each predator for a fixed trajectory in prey and predator task (Multiagent Particles environment \cite{lowe2017multi}). \textbf{Left figure}: Trajectory sampled by an expert policy. The square represents the initial position whereas the circle indicates the final position of each agent. The dots on the trajectory indicate each agent’s temporary positions. \textbf{Right figures}: normalized credit assignments generated by different multiagent RL algorithms according to this trajectory. SQDDPG presents fairer credit assignments in comparison with other methods. Reproduced with permission of Jianhong Wang \cite{wang2019shapley}.
}
\end{figure}

\subsubsection{Explanation through hierarchical goals}
 
Methods based on Hierarchical RL \cite{seijen2017hybrid} and sub-task decomposition \cite{Kawano13} consist of a high level agent dividing the main goal into sub-goals for a low-level agent, which follows them one by one to perform the high-level task. By learning what sub-goals are optimal for the low-level agent, the high-level agent forms a representation of the environment that is interpretable by humans. Often, Hindsight Experience Replay (HER) \cite{andrychowicz2017hindsight} is used in order to ignore whether or not goals and sub-goals have been reached during an episode and to extract as much information as possible from past experience.

\begin{figure}[htbp!]
    \centering
    \includegraphics[scale=0.6]{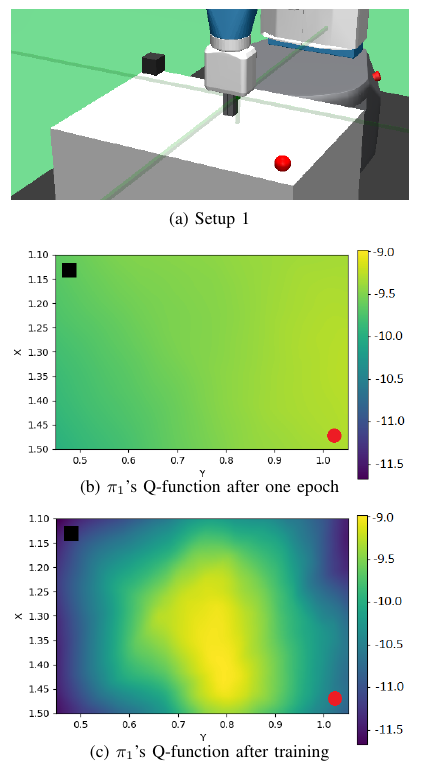}
    \caption{Setup with initial state and goal diagonally opposed on the table. The heat maps show the value of the different areas (highest in yellow) for the high-level agent to predict a sub-goal. Black squares represent the position of the cube, the red circle is the end goal. Thus, the low-level agent will have a succession of sub-goals (e.g. multiple actions that the robotic arm must perform such as moving or opening its pinch) that will ultimately lead to the achievement of the high-level goal (i.e. grasping the red ball). Reproduced with permission of \cite{Dot-to-Dot}.}
    \label{fig:dot-to-dot}
\end{figure}

Beyret et al. \cite{Dot-to-Dot} used this kind of methods along with HER for robotic manipulation (grasping and moving an item). The high level agent learns which sub-goals can make the low level agent reach the main goal while the low level agent learns to maximise the rewards for these sub-goals. The high-level agent provides a representation of the learned environment and the Q-values associated, which can be represented as heat maps as shown in Fig. \ref{fig:dot-to-dot}.

\begin{figure}[htbp!]
    \centering
    \includegraphics[scale=0.7]{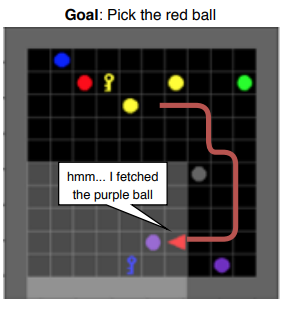}
    \caption{MiniGrid environment \cite{gym_minigrid}, where the agent is instructed through a textual string to pick up an object and place it next to another one.
    The model learns to represent the achieved goal (e.g. "Pick the purple ball") via language. As this achieved goal differs from the initial goal ("Pick the red ball"), the goal mapper relabels the episode, and both trajectories are appended to the replay buffer. Reproduced with permission of M. Seurin \cite{cideron2019selfeducated}.}
    \label{fig:selfeducated}
\end{figure}

Based on the same ideas, Cideron et al. \cite{cideron2019selfeducated} proposed Textual Hierarchical Experience Replay (THER) which extends the HER explanation to a natural language setting, allowing to learn from past experiences and to map goals to trajectories without the need of an external expert. The mapping function labels unsuccessful trajectories by automatically predicting a substitute goal. THER is composed of two models: the instruction generator which outputs a language encoding of the final state, and an agent model which picks an action given the last observations and the language-encoded goal. The model learns to encode goals and states via natural language, and thus can be interpreted by a human operator (Fig. \ref{fig:selfeducated}).

Another interesting work finds inspiration in human behaviour to improve generalization on a room navigation task, just like common  sense  and  semantic  understanding  are  used  by  humans  to  navigate  unseen environments \cite{yucommonsense}. The entire model is composed of three parts: 1) a semantically grounded navigator used to predict the next action. 2) a common sense planning module, used for route planning. It predicts the next room, based on the observed scene, helps finding intermediate targets, and learns what rooms are near the current one. 3) the semantic grounding module used to recognize rooms; it allows the detection of the current room and incorporates semantic understanding by generating questions about what the agent saw (”Did you see a bathroom?”). Self-supervision is then used for fine tuning on unseen environment. 
The explainability can be brought from the outputs of all parts of the entire model. We can get information about what room is detected by the agent, what are the next rooms targeted (sub-goals), what are the rooms predicted around the current room and what are the rooms already seen by the agent.

An original idea proposed by Tasse et al. \cite{tasse2020boolean} consists of making an agent learn basic tasks and then allow it to perform new ones by composing the tasks previously learned in a boolean formula (i.e., with conjunctions, disjunctions and negations). The main strength of this method is that the agent is able to perform new tasks without the necessity of a learning phase. From an XRL point of view, the explainability comes from the fact that the agent is able to express its actions as boolean formulas, which are easily readable by humans.  

\subsection{Post-Hoc explainability}
 
Post-Hoc explainability refers to explainability methods that rely on an analysis done after the RL algorithm finishes its training and execution. In other terms, it is a way of "enhancing" the considered RL algorithm from a black box to something that is somewhat explainable. Most Post-Hoc methods encountered were used in a perception context, i.e., when the data manipulated by the RL algorithm consisted of visual input such as images.

\subsubsection{Explanation through saliency maps}
 
When an RL algorithm is learning from images, it can be useful to know which elements of those images hold the most relevant information (i.e., the salient elements). These elements can be detected using saliency methods that produce saliency maps \cite{Selvaraju_2019, mundhenk2019efficient}. In most cases, a saliency or heat map consists of a filter applied to an image that will highlight the areas salient for the agent.

A major advantage of saliency maps is that it can produce elements that are easily interpretable by humans, even non-experts. Of course, the interpreting difficulty of a saliency map greatly depends on the saliency method used to compute that map and other parameters such as the color scheme or the highlighting technique. A disadvantage is that they are very sensitive to different input variations, and schemes to debug such visual explanation may not be straightforward \cite{jain2019attention}.

\begin{figure}[htbp!]
    \centering
    \includegraphics[scale=0.5]{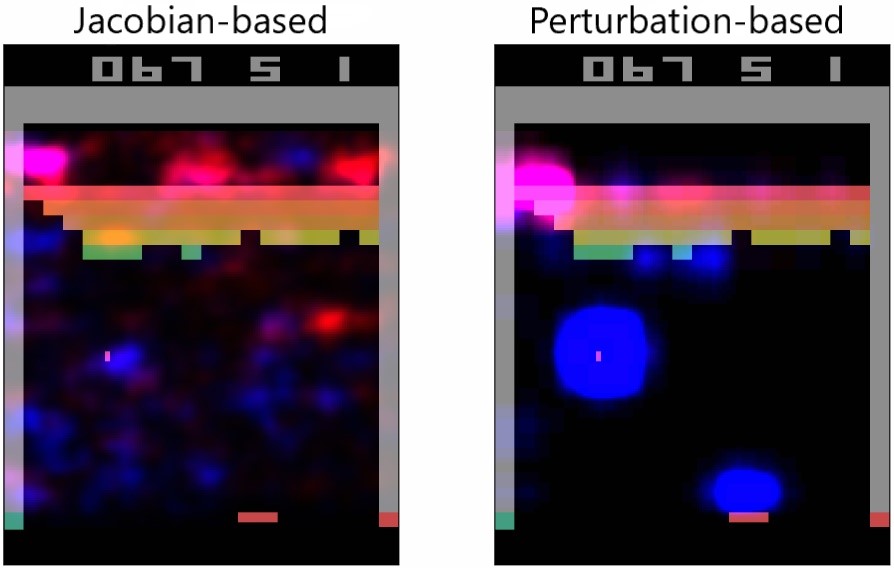}
    \caption{Comparison of Jacobian saliency (left) first introduced by Simonyan et al. \cite{simonyan2013deep} to the authors' perturbation-based approach (right) in an actor-critic model. Red indicates saliency for the critic; blue is saliency for the actor. Reproduced with permission of Sam Greydanus \cite{atarivisualizing}.}
    \label{fig:atari}
\end{figure}

A very interesting example \cite{atarivisualizing}, introduces a new perturbation-based saliency computation method that produces crisp and easily interpretable saliency maps for RL agents playing OpenAI Gym environment \textit{Atari 2600} games with Asynchronous Actor Critic \cite{mnih2016asynchronous}. The main idea is to apply a perturbation on the considered image that will remove information from a specific pixel without adding new information (by generating an interpolation from a Gaussian blur of the same image). Indeed, this perturbation can be interpreted as adding spatial uncertainty to the region around its point of application. This spatial uncertainty can help understand how removing information in a specific area of the input image affects the agent’s policy, and is quantified with a saliency metric $S$. %. The authors created a saliency metric $S$ capable to quantify this. 
The saliency map is then produced by computing $S(i,j)$ for every pixel $(i, j)$ of the input image, leading to images such as those in Fig. \ref{fig:atari}.

However, saliency methods are not a perfect solution in every situation, as pointed out in \cite{kindermans2017unreliability,adebayo2018sanity}. They need to respect a certain number of rules, such as implementation invariance or input invariance in order to be reliable, especially when it comes to their relation with either the model or the input data.

\subsubsection{Explanation through interaction data}
 
In a more generic way, the behaviour of an agent can be explained by gathering data from its interaction with the environment while running, and analysing it in order to extract key information. For instance, Caselles-Dupré et al demonstrate that symmetry-based disentangled representation learning requires interaction and not only static perception \cite{casellesdupr2019symmetrybased}.

This idea is exploited by Sequeira et al. \cite{sequeira2019interestingness2} where interaction is the %founding stone
core basis upon which their Interestingness Framework is built. This framework relies on introspection, conducted by the autonomous RL agents: the agent extracts \textit{interestingness elements} that denote meaningful interactions from their history of interaction with the environment. This is done using interaction data collected by the agent that is analysed using 
statistical methods organized in a three-level introspection analysis: 
level 0: Environment analysis, level 1: Interaction analysis; level 3: Meta-analysis. From these \textit{interestingness elements}, it is then possible to generate visual explanations (in the form of videos compiling specific highlight situations of interest in the agent's behaviour), where the different introspection levels and their interconnections provide contextualized explanations.

\begin{figure}[htbp!]
     \includegraphics[width=\textwidth]{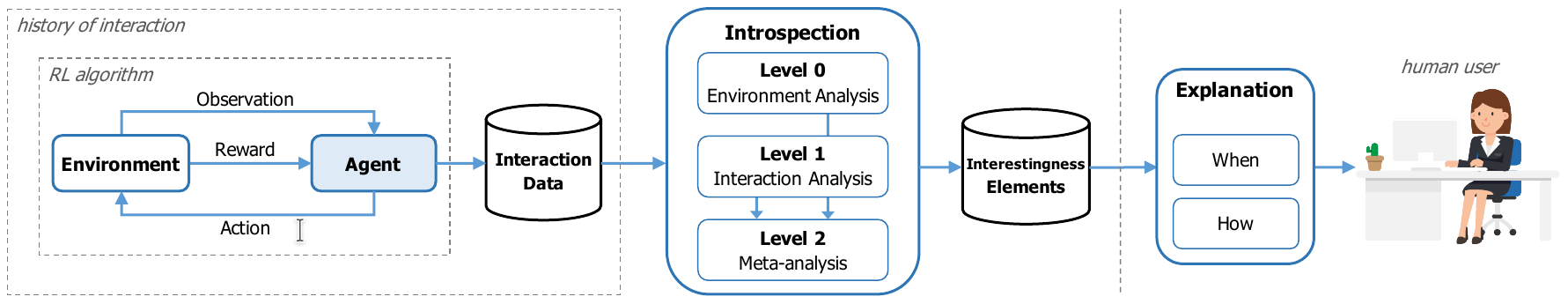}
     \caption{The interestingness framework. The introspection framework analyses interaction data collected by the agent and identifies interestingness elements of the interaction history. These elements are used by an explanation framework to expose the agent’s behavior to a human user. Reproduced with permission from Pedro Sequeira \cite{sequeira2019interestingness2}.}
     \label{fig:interestingness}
\end{figure}

The authors applied their framework to the game \textit{Frogger} and used it to generate video highlights of agents that were included in a user study. The latter showed that no summarizing technique among those used to generate highlight videos is adapted to all types of agents and scenarios. A related result is that agents having a monotonous, predictable performance will lack the variety of interactions needed by the \textit{interestingness} framework to generate pertinent explanations. Finally, counter-intuitively, highlighting all different aspects of an agent's interactions is not the best course of action ,as it may confuse users by consecutively showing the best and poorest performances of an agent.

\subsection{Other concepts aiding XRL}
 
Some studies encountered do not fit in the above categories for the main reason that they are not linked to RL or do not directly provide explanations but nonetheless, they are interesting concepts that could contribute to the creation of new XRL methods in the future.

\subsubsection{Explainability of CNNs}
 
Although deep neural networks have exhibited superior performance in various tasks, their interpretability is always their Achilles’ heel. Since CNNs are still considered black boxes, many recent research papers focus on providing different levels and notions of explanations to make them more explainable. 
 
As many RL models harness visual input DL models (for instance, when processing pixel observations), they could profit from better explainability of these algorithms. That way, the complete block of a CNN associated to learn a policy, % will be explainable as a whole. 
would be explainable as whole. In addition, some techniques used in the visual domain, such as representation disentanglement could be relevant to apply in RL. Among the approaches detailed by Zhang et al. \cite{zhang2018visual}, one of the most promising aims at creating disentangled (interpretable) representations of the conv-layers of these networks \cite{zhang2016growing, zhang2017interpreting}, as well as end-to-end learning of interpretable networks, working directly with comprehensible patterns, which are also a trending angle  \cite{wu2017interpretable}.

Explaining when, how, and under which conditions catastrophic forgetting \cite{dazrodrguez2018dont} or memorizing of datasets occurs is another relevant aspect of life-long or continual learning \cite{Lesort2019ContinualLF} in DNNs yet not fully understood. 
An interesting method towards this vision is Learning Without Memorizing (LwM) \cite{dhar2018learning}, an extension of Learning Without Forgetting Multi-Class (LwF-MC) \cite{li2016learning} applied to image classification. This model is able to incrementally learn new classes without forgetting classes previously learned and without storing data related them. The main idea is that at each step, a new model, the student, is trained to incrementally learn new classes, while the previous one, the teacher, only has knowledge of the base classes. By improving LwF-MC with the application of a new loss called Attention Distillation loss, LwM tries to preserve base classes knowledge across all models iterations. This new loss produces attention maps that can be studied by a human expert in order to interpret the model's logic by inspecting the areas that focus its attention.

Another approach for scene analysis aimed to build a graph where each node represents an object detected in the scene and is capable of building a context-aware representation of itself by sending messages to the other nodes \cite{hu2019languageconditioned}. This makes it possible for the network to support relational reasoning, allowing it to be effectively transparent. Thus, users are able to make textual inquiries about relationships between objects (e.g., "Is the plate next to a white bowl?"). 

\subsubsection{Compositionality as a proxy tool to improve understandability}

Compositionality is a universal concept stating that a complex (composed) problem can be decomposed into a set of simpler ones \cite{Baroni_2019}. Thus, in the RL world, this idea can be translated into making an agent solve a complex task by hierarchically completing lesser ones (e.g. by first solving atomic ones as lesser tasks could also be complex) \cite{pierrot2019learning}. This provides reusability, enables quick initialization of policies  and makes the learning  process much faster by training  an optimal policy for each reward and later combining them. Haarnoja et al. \cite{Haarnoja_2018} showed that maximum entropy RL methods can produce much more composable policies. Empirical demonstrations were performed on a Sawyer robot trained to avoid a fix obstacle and to stack Lego blocks with both policies combined. They introduced the Soft Q-learning algorithm, based on maximum entropy RL \cite{ziebart2008maximum} 
and energy-based models \cite{haarnoja2017reinforcement}, as well as an extension of this algorithm that enables composition of learned skills. This kind of methods optimizing for compositionality does not provide a direct explanation tool; however compositionality can be qualitatively observed as self organized modules \cite{Han19} and used to train multiple policies that benefit from being combined. Compositionality may also help better explain each policy along the training evolution in time, or each learned skill separately. However, it is also observed that compositionality may not emerge in the same manner as humans conceptually would understand it or expect it, e.g. based on symbolic abstract functionality modules. Some examples in language emergence in multi-agent RL settings show that generalization and acquisition speed \cite{kharitonov2020emergent} or language do not co-occur with compositionality, or that compositionality may not go hand in hand with language efficiency as in humans communication \cite{Chaabouni19}.

Distillation has also been used to learn task that are closely related and whose learning should improve speed up the learning of near tasks, in DisCoRL model \cite{traore2019discorl}, which helps transfer from simulation to real settings in navigation and goal based robotic tasks. We may then be able to further explain  each policy along the training evolution timeline, or each learned skill separately.

\subsubsection{Improving trust via imitation learning}
 
Imitation learning is a way of enabling algorithms to learn from human demonstrations, such as teaching robots to learn assembly skills \cite{10.3389/frobt.2019.00099, Ng2004}. While improving training time (compared to more traditional approaches \cite{doncieuxSRL2018}), this method also allows for better understanding of the agent's behaviour as it learns according to human expert actions \cite{christiano2017deep}.
It can also be a way to improve trust in the model, as it behaves seemingly as a human expert operator and can explain the basis of its decisions textually or verbally. Moreover, when encompassing human advice during training, it can be derived into advisable learning which further improves user trust as the model can understand human natural language and yields clear and precise explanations \cite{Kim_2020_CVPR}.

\subsubsection{Transparency-oriented explanation building}
 
Transparency has been given multiple meanings over time, especially in robotics and AI Ethics. Theodorou et al. \cite{10.1080/09540091.2017.1310182} freshly define it as a mechanism to expose decision making that could allow AI models to be debugged like traditional programs, as they will communicate information about their operation in real time. However, the relevance of this information should adapt to the user's technological background, from simple progress bars to complex debug logs. An interesting concept is that an AI system could be created using a visual editor that can help communicate which decision will be taken in which situation (very much like decision trees). These concepts have already been successfully implemented in an RL setup using Temporal Difference (TD) error to create an emotional model of an agent \cite{matarese2020transparency}.

\section{Discussion } \label{sec:discussion}

Despite explainable deep RL being still an emerging research field, we observed that numerous approaches were developed so far, as detailed in
\hyperref[sec:2]{Section 2}. However, there is no clear-cut method that serves all purposes. Most of the reviewed XRL methods are specifically designed to fit a particular task, often related to games or robotics and with no straight forward extension to other real-world RL applications. Furthermore, those methods cannot be generalized to other tasks or algorithms as they often make specific assumptions (e.g. on the MDP or environment properties). In fact in XRL there can be more than one model (as in Actor-Critic architectures) and different kinds of algorithms (DQN, DDPG, SARSA...) each with its own particularities. Moreover, there exists a wide variety of environments where each brings its own constraints. The necessity to adapt to the considered algorithm and environment means that it is hard to provide a %really 
holistic or \textit{generic} explainability method. Thus, in our opinion, Shapley value-based methods \cite{lundberg2017unified, wang2019shapley} can be considered as an interesting lead to contribute to this goal. Shapley values could be used to explain the roles taken by agents when learning a policy to achieve a collaborative task but also to detect defects in training agents or in the data fed to the network. In addition, as a post-hoc explainability method, it may be possible to generalize Shapley value computation to numerous RL environments and models in the same way it was done with SHAP \cite{lundberg2017unified} for other black boxes Deep Learning classifiers or regressors.

Meanwhile, the research community would benefit if more global-oriented approaches, which do not focus on a particular task or algorithm, were developed in the future, as it has already been done in general XAI, with for instance LIME \cite{ribeiro2016i} or SHAP \cite{lundberg2017unified}.

Moreover, some promising approaches to bring explainability to RL include representation learning related concepts such as Hindsight Experience Replay, Hierarchical RL and self-attention. However, despite the ability of those concepts to improve performance and interpretability in a mathematical sense (in particular representation learning), they somehow lack concrete explanations targeted to end users, as they mostly target technical domain experts and researchers.  This is a key element to further develop and allow the deployment of RL in the real world and to make algorithms more trustable and understandable by the general public.

The state of the art shows there is still room for progress to be made to better explain deep RL models in terms of different invariants preservation and other common assumptions of disentangled representation learning \cite{locatello2018challenging,achille2017separation}.

\section{Conclusion and Future Work} \label{sec:conclusion}

We reviewed and analyzed different state of the art approaches on RL and how XAI techniques can elucidate and inform their training, debugging and communication to different stakeholder audiences. 

We focused on agent based RL in this work, however, explainability in RL involving humans (e.g. in collaborative problem solving \cite{bennetot2020should}) should involve explainability methods to better assess when robots are able to perform the requested task, and when uncertainty is an indicator of better relying a task to a human. 
Equally important is to evaluate and explain other aspects in reinforcement learning, e.g. formally explaining the role of curriculum learning \cite{portelas2020automatic}, quality diversity or other human-learning inspired aspects of open-ended learning \cite{doncieux2020dream, mouret2015illuminating, pugh2016quality}.
Thus, more theoretic bases to serve explainable by design DRL are required. The future development of post-hoc XAI techniques should adapt to the requirements to build, train, and convey DRL models. Furthermore, it is worth noting that all presented methods decompose final prediction into additive components attributed to particular features \cite{Staniak18}, and thus interaction between features should be accounted for, and included in the explanation elaboration. Since most presented strategies to explain RL have mainly considered discrete model interpretations for explaining a model, as advocated in \cite{Sundararajan20}, continuous formulations of the proposed approaches (such as Integrated Gradients \cite{Sundararajan17} based on the continuous extension of Shapley value, Aumann-Shapley value cost-sharing technique) should be devised in the future in RL contexts.

We believe the reviewed approaches and future extensions tackling the identified issues will likely be critical in the demanding future applications of RL. We advocate for the needs of targeting in the future more diverse audiences (developer, tester, end-user, general public) not yet approached in the development of XAI tools. Only this way we will produce actionable explanations and more comprehensive frameworks for explainable, trustable and responsible RL that can be deployed in practice.

\section*{Acknowledgements}

We thank Sam Greydanus, Zoe Juozapaitis, Benjamin Beyret, Prashan Madumal, Pedro Sequiera, Jianhong Wang, Mathieu Seurin and Vinicius Zambaldi for %agreeing 
allowing us to use their original images for illustration purposes.
We also would like to thank Frédéric Herbreteau and Adrien Bennetot for their help and support.

\newpage

\bibliography{refs}

\newpage

\section{Appendix}
\label{ref:appendix}
\subsection{Glossary}
\begin{itemize}
    \item \textbf{A2C}: Asynchronous Actor Critic \cite{mnih2016asynchronous}
    \item \textbf{AI}: Artificial Intelligence
    \item \textbf{COMA}: Counterfactual multi-agent \cite{foerster2017counterfactual}
    \item \textbf{CNN}: Convolutional Neural Network \cite{lecun1995convolutional}
    \item \textbf{DDPG}: Deep Deterministic Policy Gradient \cite{lillicrap2015continuous}
    \item \textbf{DL}: Deep Learning
    \item \textbf{DRL}: Deep Reinforcement Learning
    \item \textbf{DQN}: Deep Q Network \cite{mnih2013playing}
    \item \textbf{GAN}: Generative Adversarial Network  \cite{goodfellow2014generative}
    \item \textbf{HER}: Hindsight Experience Replay \cite{andrychowicz2017hindsight}
    \item \textbf{HMM}: Hidden Markov Model
    \item \textbf{HRA}: Hybrid Reward Architecture \cite{seijen2017hybrid}
    \item \textbf{HRL}: Hierarchical Reinforcement Learning \cite{Kawano13}
    \item \textbf{IDDPG}: Independent DDPG  \cite{lillicrap2015continuous}
    \item \textbf{MADDPG}: Multiagent DDPG  \cite{lowe2017multi}
    \item \textbf{MDP}: Markov Decision Process 
    \item \textbf{Machine Learning}: Machine Learning
    \item \textbf{POMDP}: Partialy Observable Markov Decision Process
    \item \textbf{PPO}: Proximal Policy Optimization \cite{schulman2017proximal}
    \item \textbf{R-CNN}: Region Convolutionnal Neural Network \cite{rcnn2013}
    \item \textbf{RL}: Reinforcement Learning 
    \item \textbf{SARSA}: State Action Reward State Action \cite{sarsa}
    \item \textbf{SRL}: State Representation Learning \cite{lesort2018srlOverview}
    \item \textbf{VAE}: Variational Auto-Encoder \cite{kingma2013autoencoding}
    \item \textbf{XAI}: Explainable Artificial Intelligence
    \item \textbf{XRL}: Explainable Reinforcement Learning
\end{itemize} 

\medskip

\end{document}